\documentclass[runningheads]{llncs}
\usepackage{graphicx}
\usepackage{multirow}
\usepackage{booktabs}
\usepackage{amsmath}
\usepackage{cite}
\begin{document}
\title{Comparison of Machine Learning Models to Classify Documents on Digital Development}
\titlerunning{ML Models to Classify Digital Development Reports}
\author{Uvini Ranaweera\inst{1}\orcidID{0009-0003-6205-3482} \and
Bawun Mawitagama\inst{1}\orcidID{0009-0008-2893-0097} \and
Sanduni Liyanage\inst{1}\orcidID{0009-0004-2864-7292}\and
Sandupa Keshan\inst{1}\orcidID{0009-0009-0924-9711} \and
Tiloka De. Silva\inst{1}\orcidID{0000-0002-0543-2182} \and
Supun Hewawalpita\inst{1}\orcidID{0009-0000-4787-8468}}
\authorrunning{U. Ranaweera et al.}
\institute{University of Moratuwa, Katubedda 10440, Sri Lanka \\ 
\email{\{ranaweeraraua.19, mawitagamabc.19, liyanagesc.19, keshanams.19, tilokad, supungs\}@uom.lk}}

\maketitle              
\vspace{0.1cm}
This version has been accepted for publication after peer review, but is not the Version of Record. The Version of Record is available at \url{https://doi.org/10.1007/978-981-99-7969-1\_5}. Use of this Accepted Version is subject to Springer’s terms of use.
\begin{abstract}
Automated document classification is a trending topic in Natural Language Processing (NLP) due to the extensive growth in digital databases. However, a model that fits well for a specific classification task might perform weakly for another dataset due to differences in the context. Thus, training and evaluating several models is necessary to optimise the results.  This study employs a publicly available document database on worldwide digital development interventions categorised under twelve areas. Since digital interventions are still emerging, utilising NLP in the field is relatively new. Given the exponential growth of digital interventions, this research has a vast scope for improving how digital-development-oriented organisations report their work.  The paper examines the classification performance of Machine Learning (ML) algorithms, including Decision Trees, k-Nearest Neighbors, Support Vector Machine, AdaBoost, Stochastic Gradient Descent, Naive Bayes, and Logistic Regression. Accuracy, precision, recall and F1-score are utilised to evaluate the performance of these models, while oversampling is used to address the class-imbalanced nature of the dataset. Deviating from the traditional approach of fitting a single model for multiclass classification, this paper investigates the One vs Rest approach to build a combined model that optimises the performance. The study concludes that the amount of data is not the sole factor affecting the performance; features like similarity within classes and dissimilarity among classes are also crucial.

\end{abstract}

\section{Introduction}

With the increased integration of technology in operations, electronic documents or e-documents have seen significant prominence. E-documents usually include Portable Document Format (PDF), text files, e-mails, HTML, and PostScript~\cite{caldas2002automated}. The use of such documents has grown due to properties like accessibility and convenience pushing towards the information explosion era~\cite{hakim2014automated}. Because of the growing use of e-documents, databases are developed with predefined categories, and these documents are classified based on their content. This classification enhances the document management mechanism within the database and makes it convenient for users to locate and refer to the documents based on their specific requirements. 

In most databases, the categories of documents are determined by human annotators. However, given the exponential increase in e-documents available, manual classification has become a strenuous task and continuing the process has been challenging. Though it is time-consuming, the task is an easy decision for human beings to make~\cite{borko1963automatic}. As a result of continuous experiments, e-document classification has now been automated with user-defined categories and is widely used for databases~\cite{borko1963automatic}. Automated document classification uses computer programs to map documents into one or more predefined classes~\cite{cohen2006effective}. Many attempts have been introduced in the literature to classify domain-specific documents~\cite{behera2019performance,pandey2017automated,wei2018empirical,zhang2019construction}. 

Along the line, digital development initiatives are a rising domain to be embedded in Knowledge Management Systems. The field of digital development has seen a leap in its significance in the world with digital transformation and the growing reliance on digital technology. This surge has led to increased reports on the topic, broadly covering areas like digital infrastructure, literacy, finance, services, data systems and education. Effective classification of documents in the digital development field is important as it facilitates stakeholders to engage in knowledge sharing, informed decision making and targeted analysis with the help of insights, trends and other useful information extracted from these documents. Since the field is still expanding, the documents available are limited, making it possible for human annotators to classify the available documents manually. As a result, the world today has many documents accumulated on databases specified for digital development. This does not translate to millions of reports but several thousands of digital development reports accessible online. However, over time, the reports on digital development will grow significantly, making the manual classification process challenging. 

This paper presents the results of automated document classification in the digital development domain. To address all the domain-specific issues mentioned, the study employs a dataset with twelve predefined classes and is class imbalanced due to fewer data points. A range of ML algorithms like Decision Trees, k-Nearest Neighbors (k-NN), Support Vector Machine (SVM), AdaBoost, Stochastic Gradient Descent (SGD), Naive Bayes and Logistic Regression~\cite{hakim2014automated} are employed in classifying the documents.\footnote{At the beginning of the study, the performance of Deep Learning (DL) algorithms such as Convolutional Neural Networks (CNNs), Recurrent Neural Networks (RNNs) and Transformers were measured. However, training DL algorithms was challenging, given the limitations in computational resources and the number of data points available. With low performance recorded, it was unanimously decided to drop the DL algorithms and proceed with the ML algorithms for this study.}

Another unique aspect of this study is that the dataset used is not specifically designed for a research purpose and exhibits thousands of features per record, increasing the practicality of the research. 

Since going ahead with a single ML model to perform multiclass classification on a limited number of documents is challenging, the study adopts the One vs Rest (OvR) approach, assuming that the same model might not perform equally well for all the classes. It identifies the best classifier for each class individually, after which an overall model is fitted by combining the classifiers. The dataset being class-imbalanced (the number of records among categories is unequal) is another challenge to handle, which could result in inaccurate classifications if left unaddressed~\cite{hakim2014automated}. 

The rest of the research paper is structured as follows. Section 2 reviews the literature on text document classification using various ML algorithms. Section 3 describes the methodology, which includes an outline of the conceptual framework followed in the study, an introduction to the dataset, the preprocessing techniques applied, the two phases of the distinct models developed for classification, and the performance measures used for model evaluation.  Section 4 analyses the results and carries out an extensive discussion of the findings of the study. Section 5 gives the conclusion at the end of the study and suggestions for future works to enhance the functionality of the research.

\section{Literature Review}
Natural Language Processing (NLP) has a wide range of applications that improve word processing significantly. Church and Rau~\cite{church1995commercial} suggested that NLP includes advanced data organisation features such as spelling correction, dictionary access, and categorisation. Furthermore, they stated that NLP improves information management and retrieval systems by automating data classification and extraction. NLP contributes significantly to accurate and organised content creation by improving word processing capabilities. Sebastiani~\cite{sebastiani2002machine} showed that automated document classification classifies text documents using computational techniques into predefined categories. It entails labelling or tagging documents according to their content using tools in ML and NLP. 

Many studies have been carried out worldwide to evaluate and suggest methodologies to automate the process of text classification. Kowsari et al.~\cite{kowsari2019text} have surveyed algorithms for text classification. The study concluded that algorithms such as Rocchio, boosting, bagging, logistic regression, Naïve Bayes, k-NN and SVM have their own pros and cons in text classification. Ting et al.~\cite{ting2011naive} examined Naive Bayes' suitability as a document classifier. The findings demonstrated that Naive Bayes outperforms other classifiers in precision, recall, and F-measure, achieving a high classification accuracy. The performances of the k-NN and Naive Bayes algorithms for classifying text documents were compared by Rasjid and Setiawan~\cite{rasjid2017performance} using the TREC Legal Track dataset. The study concluded that Naive Bayes outperforms k-NN with k=1 in terms of recall. However, as k rises, k-NN outperforms Naive Bayes in terms of F-measure, recall, and precision. k = 13 is found to be the ideal value for the dataset, producing stable and precise results. The lack of specific content in the text documents contributes to the overall poor performance, but the study suggested using k-NN with k values greater than 10 for better classification. 

Moreover, studies have been carried out in the recent past on domain-specific document classification. Wei et al.~\cite{wei2018empirical} focused on applying DL algorithms such as CNN in the legal field, where large sets of electronically saved data are available. The study found that ML algorithms like SVM are generally suited for smaller datasets, whereas CNN performs significantly for larger datasets.  Pandey et al.~\cite{pandey2017automated} talked about how ML algorithms like Naïve Bayes, Linear Discriminant Analysis (LDA), K-NN, SVM with different kernels, Decision Trees and Random Forests can be used to classify issue reports in the field of software development. Random Forests and SVM with certain kernels performed well under the metrics F-measure, Average accuracy, and weighted F-measure on 7401 issue reports extracted from five open-source projects; three of these projects were tracked through Jira and two through Bugzilla. Behera et al.~\cite{behera2019performance} described how DL algorithms CNN, RNN, Deep Neural Network (DNN) and Ensemble Deep Learning models are used in the biomedical domain and how the performance of these models outweighs that of ML algorithms. Using three datasets: Ads dataset, TREC 2006 Genomics Track dataset and BioCreative Corpus III (BC3), it was found that, with the increase of the dataset size, the use of DL algorithms is much more successful than ML algorithms. Zhang et al.~\cite{zhang2019construction} discussed how ML algorithms such as SVM, K-NN, Linear regression, Decision Trees, Naïve Bayes, and ensemble methods could analyse reports on construction site accidents. The weighted F1 score indicated that the optimised ensemble method performs the best.

As stated above, though there is much literature on diverse fields, less work can be found on classifying digital development-related reports. Further, it is rare to find research conducted for an imbalanced dataset, even though it is less likely to find a perfectly balanced dataset in reality.

\section{Methodology}
\subsection{Defining the Dataset and Preprocessing}

\begin{figure}[ht]
\includegraphics[width=\textwidth]{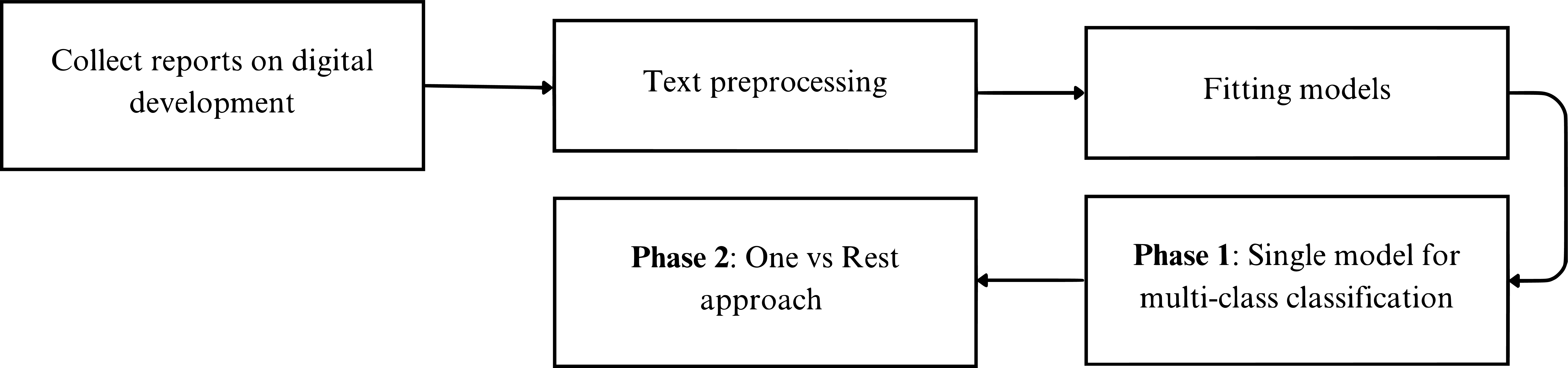}
\caption{Conceptual framework.} \label{Fig-1}
\end{figure}

\begin{figure}[ht]
\includegraphics[width=\textwidth,height=0.48\textwidth]{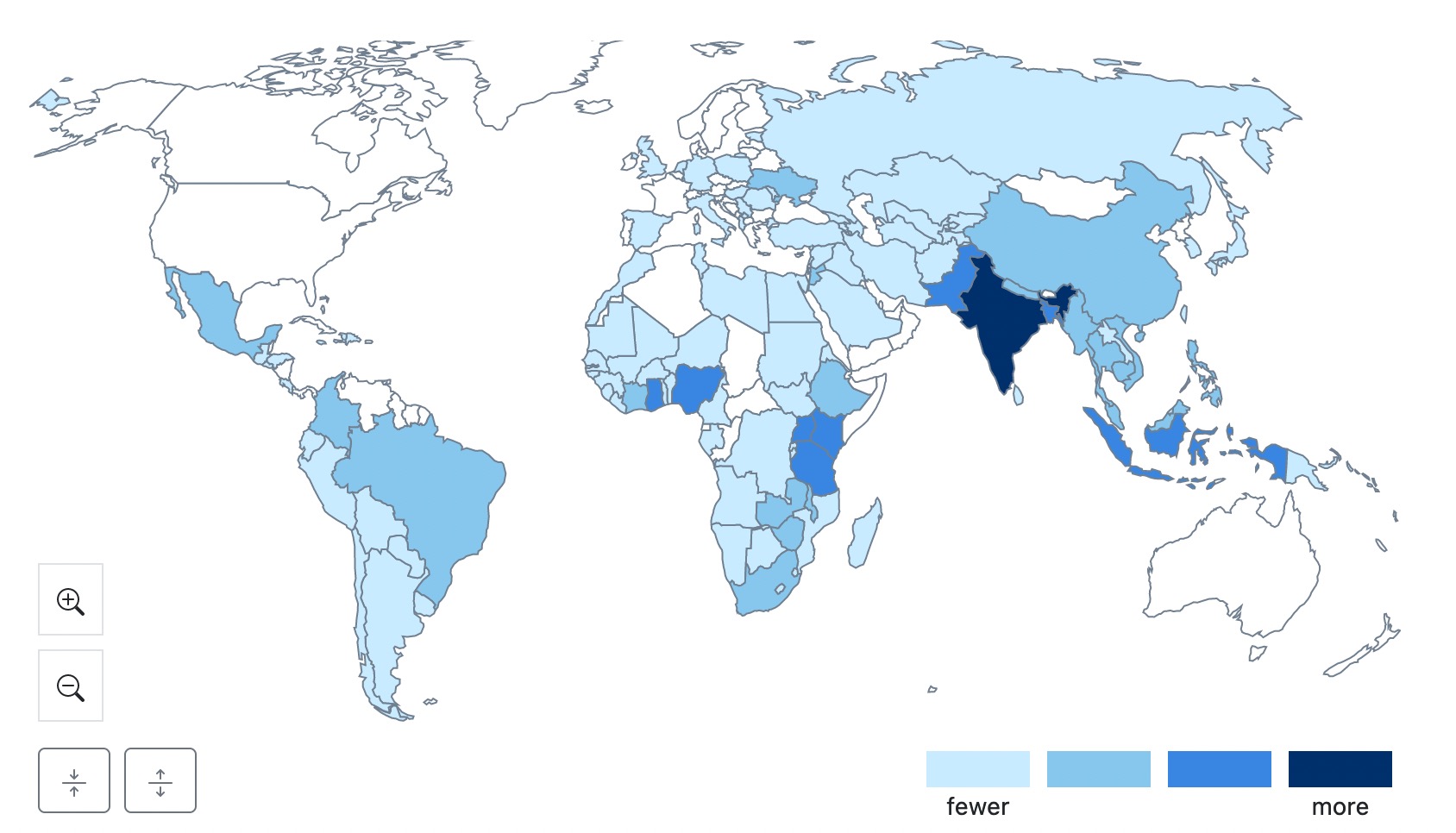}
\caption{Worldwide distribution of reports on digital development interventions.} \label{Fig-2}
\end{figure}

As indicated in Fig.~\ref{Fig-1}, a conceptual framework was designed to address the research aim of optimising the model performance in digital development report classification. Accordingly, to carry out the study, first, it is necessary to get a dataset centred around digital development initiatives. The United States Agency for International Development - Digital Ecosystem Evidence Map (USAID DEEM)~\cite{deem-website} was chosen as the primary data source to study as it is a searchable database that stores worldwide digital development evidence, making it a world map for such data (Fig.~\ref{Fig-2}).

The USAID DEEM is an evidence map compiled by the United States Agency for International Development (USAID) as a part of its digital strategy. USAID continually adds reports to this evidence map to cover the worldwide efforts in digital development interventions. As indicated in Fig.~\ref{Fig-2}, the process is partially completed to date, and still, there are only a few reports from most parts of the world.

\begin{figure}[ht]
\includegraphics[width=\textwidth]{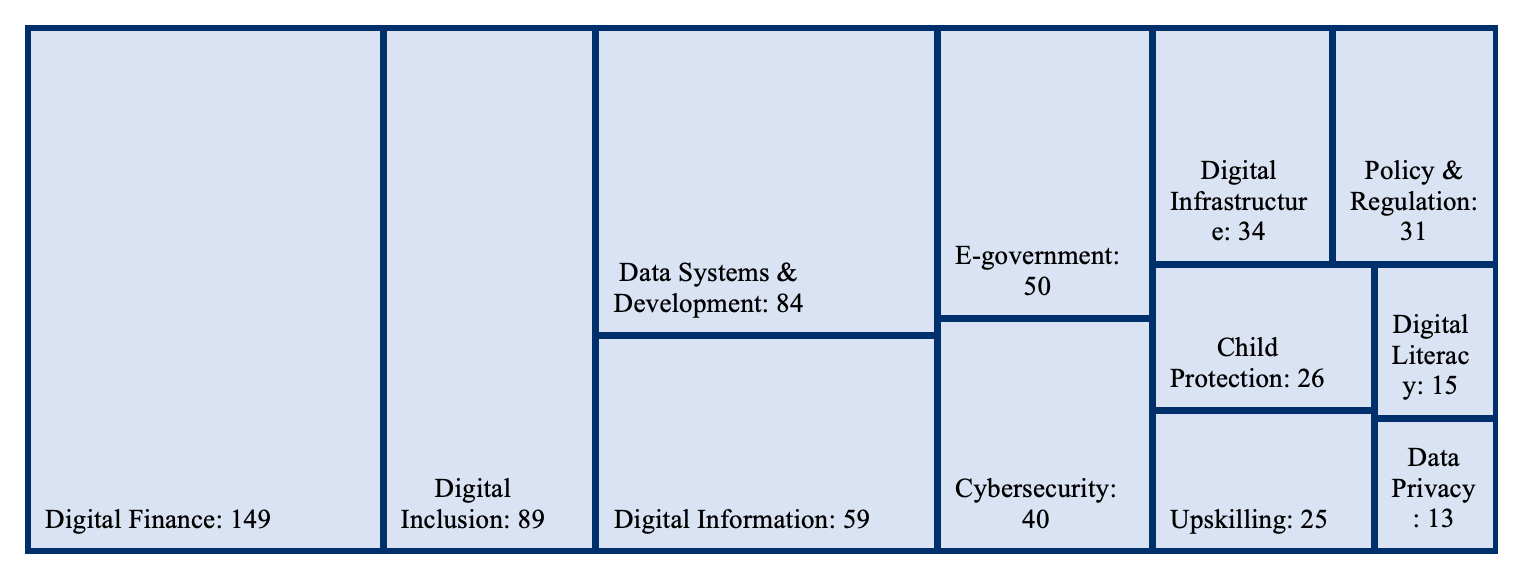}
\caption{Distribution of documents across intervention areas after removing duplicates.} \label{Fig-3}
\end{figure}

\noindent The USAID DEEM dataset includes 851 documents \footnote{Since there is no intention to train a multi-label classification model for this study, duplications were eliminated, downsizing the dataset to 615 records.} classified under twelve predefined intervention areas: Child Protection, Cybersecurity, Data Privacy, Data Systems \& development, Digital Finance, Digital Inclusion, Digital Information Services, Digital Infrastructure Development, Digital literacy, Policy \& regulation or Digital Services, E-government and Upskilling/ Capacity Building. The document database includes a variety of reports, such as research papers, project reports and case studies. However, a heavy class imbalanced nature is observed among the twelve classes, as shown in Fig.~\ref{Fig-3}.

\noindent Several factors were considered when choosing USAID DEEM as the custom dataset for this study. Since the study intends to evaluate classification models for reports on digital development, a dataset centring around the field was a must. Much research has been conducted on balanced datasets, making accuracy a more reliable estimator to measure the success of the classification. However, when gathering data from an area that is still evolving, it is impossible to expect a well-balanced dataset. So, to reflect practicality, it was necessary to have a dataset that is being used at an organisational level. 

Furthermore, most studies in text classification have utilised datasets designed for research purposes, such as newsgroups and Reuters, which include only a few features to make the process easier and faster. USAID DEEM, where the document length ranged from a few words to 100,000+ words, enriched the research to handle datasets with many characteristics. Hence, this study employs the full content of a document rather than selecting a section of an article like the heading, abstract or introduction. The approach allows us to feed a 360-degree view of the documents, enhancing model performance, given that document classification is quite a new topic for digital development reports.

 After defining the dataset, as for the conceptual framework in Fig.~\ref{Fig-1}, next up was to preprocess the text data. Text preprocessing is the first step in text mining; that needs to be done following web scraping. Since all the documents from USAID DEEM were extracted in PDF format, converting them into text format was necessary before applying any preprocessing steps. As Diem et al.~\cite{diem2011text} suggested, the Optical Character Recognition (OCR) technology was used to read each document in the database to the text format. Finally, to transfer the text documents into a dataframe, the PySpark library was utilised~\cite{singh2019natural}.

Under text preprocessing, a series of steps such as lowercasing, tokenisation, stop words removal and lemmatisation was carried out~\cite{kadhim2018evaluation} using the Python library NLTK. NLTK is an NLP toolkit that is used for most of the preprocessing steps, and Hardeniya~\cite{hardeniya2015nltk} stated it as one of the most popular libraries for NLP as it was built on the learning curve of Python, making it fast and easy to work with. It is noteworthy to mention that a domain-specific stopword list was defined for successful execution. When defining the stop word list, words like “digital”, “development”, “project”, etc., were included as they frequently occur in digital development reports, adding no value. Finally, to make the text corpus ready to be fed into ML algorithms, each text record was converted into a vector format using Term Frequency - Inverse Document Frequency (TF-IDF) vectorisation~\cite{aizawa2003information}.

For the experiment, the final dataset of 615 records was utilised, with 70\% of the data points considered as training data and the remainder as test data points. The 185 test data points were used to evaluate how well the models generalise on unseen data.

\subsection{Defining Models}
The classification algorithms were applied in two major phases, as mentioned in Fig.~\ref{Fig-1}. However, before directly fitting a model, it was necessary to treat the class-imbalanced nature of the dataset as the unequal distribution of data points among the classes may result in poor performance in minority classes and model bias for the majority classes~\cite{adil2021solving}. To further elaborate on minority and majority classes, in the USAID DEEM dataset, classes like Digital Finance and Digital Inclusion can be categorised as majority classes, while classes like Data Privacy and Digital Literacy come under minority classes. 

Among the techniques to address the class-imbalanced nature of the dataset are oversampling, undersampling and adjusting class weights. Oversampling is where new data points are added to the minority classes either by adding new instances or by repeating existing instances~\cite{mohammed2020machine}. Undersampling is where data points are removed from the majority classes~\cite{mohammed2020machine} while the class weights method will assign higher weights for minority classes and vice versa to avoid the model bias resulting from imbalanced data~\cite{yu2022re}. For the underlying study, the oversampling technique was employed since the training dataset was limited to 430 data points.

The first phase of the analysis focused on training single models capable of multiclass classification on the USAID DEEM dataset. The ML algorithms trained were Decision trees, Naïve Bayes, k-NN, logistic regression, AdaBoost, SGD and SVM. These algorithms were picked in a way to represent different aspects of ML models, such as lazy learning (k-NN), eager learning (decision trees, SVM), discriminative models (logistic regression) and generative models (Naïve Bayes). Moreover, an ensemble learning approach was adapted to evaluate classification performance by combining the two best-performing ML algorithms based on the F1 score. The ensembling process used the majority voting mechanism to decide the class from the two base models ~\cite{dong2020survey}. 

Moving on to the study’s second phase, the aim was to train the above-mentioned ML algorithms for binary classification on a class-imbalanced dataset. Given that the study deals with a multiclass dataset, to try out binary classification, a One vs Rest (OvR) approach was utilised. There were two assumptions behind adopting OvR to this study: deploying a single model to predict class labels might not perform better when the dataset is class imbalanced, and the uniqueness inherent to each class will make an algorithm record different performances among classes. Thus, instead of training one classification model for all the classes, comparing model performances and fitting the best algorithm for each class will generate better results in classification. 

Though there is no motive to try multi-label classification at this stage, the OvR approach eases addressing multi-label classification as well (one document belonging to more than one class), which is the reality for this dataset. So, trying out OvR will allow the study to expand its scope to accommodate multi-label classification. At the moment, by applying an OvR approach, this study trains individual models that perform significantly better for at least a few of the classes and tries to understand what hinders the model performance for the rest of the classes.

\subsection{Performance Measures}
Several performance measurements, such as accuracy (correctly classified instances over total instances), precision (correctly classified positive instances over total positive instances predicted), recall (correctly classified positive instances over actual positive instances), and F1(harmonic mean of precision and recall), were used to measure the performance of each algorithm~\cite{forman2003extensive}. All these performance measures can be derived using the four features: True Positive (TP), True Negative (TN), False Positive (FP), and False Negative (FN)~\cite{behera2019performance}, as summarised in Table~\ref{tab1}.
\begin{table}[ht]
\centering
\caption{Confusion matrix}\label{tab1}
\begin{tabular}{@{}lll@{}}
\toprule
& \multicolumn{2}{c}{Predicted Label} \\
\cmidrule{2-3}
Actual Label & Positive & Negative \\
\midrule
Positive & TP & FN \\
Negative & FP & TN \\ 
\bottomrule
\end{tabular}
\end{table}

\noindent Since the study deals with an imbalanced dataset, accuracy is unreliable. As an unbiased estimator, F1-score (\ref{eq:F1}), a combination of precision and recall~\cite{wardhani2019cross} can be used.

\begin{equation}
F_1 = \frac{2 \times P \times R}{P + R} \label{eq:F1}
\end{equation}

\noindent The study's phase one performance was measured using the weighted average F1-score (\ref{eq:WF1}) to account for the class-imbalanced nature under multiclass classification~\cite{behera2019performance}. 

\begin{equation}
\text{Weighted\ average\ F1} = \frac{\sum_{i=1}^{m} |y_i| \times \frac{2 \times \text{tp}_i}{2 \times \text{tp}_i + \text{fp}_i + \text{fn}_i}}{\sum_{i=1}^{m} |y_i|} \label{eq:WF1}
\end{equation}

\section{Results and Discussion}
This section of the paper investigates the results of the models trained under phase one (training single models for multiclass classification) and phase two (OvR approach for binary classification). All these models have been executed using Google CoLab Jupyter Notebook with Python 3.10 version with a maximum RAM of 12 GB.

\begin{figure}[ht]
\centering
\includegraphics[width=0.7\textwidth,height=0.5\textwidth]{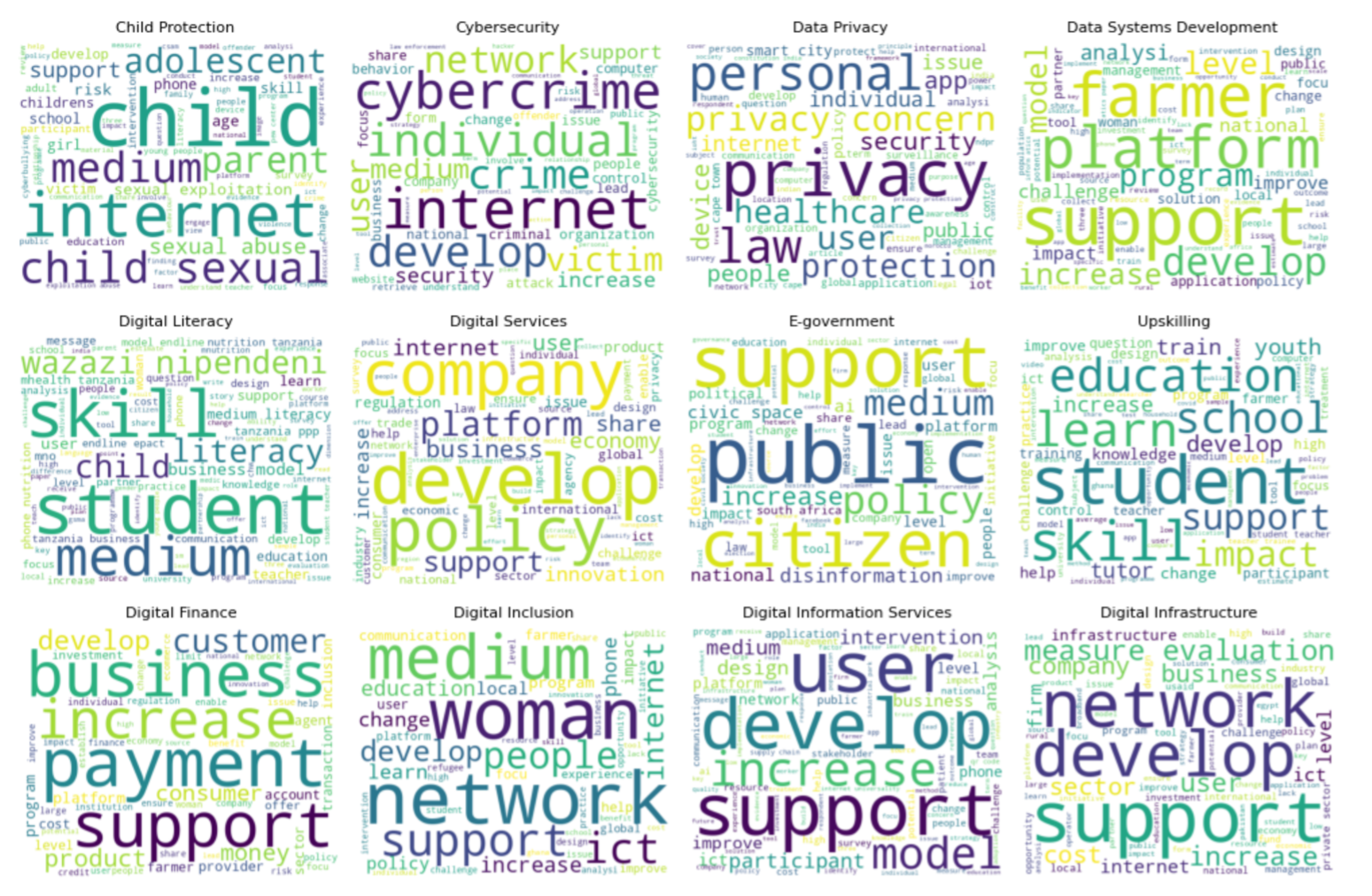}
\caption{Word clouds for intervention areas.} \label{Fig-4}
\end{figure}

First, an Exploratory Data Analysis (EDA) was carried out to get insights into the dataset.  As for Fig.~\ref{Fig-4}, the twelve distinct classes of the dataset have similarities and dissimilarities, making it an interesting point to study how the models would perform classification.

\noindent Before fitting the models under the first phase, the class labels, which took categorical values, were encoded numerically. Moreover, since the dataset was heavily class imbalanced, random oversampling was applied, which balanced the training dataset with each class having exactly 100 records. In line with the class-imbalanced nature of the dataset, the weighted average F-1 score was used over accuracy as an unbiased measure of model performances on the test dataset. 

From the weighted average F-1 scores recorded in Table~\ref{tab2}, it is evident that logistic regression outperformed the rest with an F1-score of 0.53, followed by SGD with a score of 0.51. It is a noticeable fact that AdaBoost has recorded the worst results for phase one. Even literature has strong evidence for AdaBoost reporting worst performances in text classification~\cite{al2019arabic,gutman2015text}.

As per literature~\cite{lipton2014optimal}, F1- measures should be optimised to generalise a model for further classifications. The results indicated in Table~\ref{tab2} show that none of the models succeeds in going beyond 0.53, and most of them fall under a mid-range performance (except for Decision trees and AdaBoost, all the other classifiers have reported a performance score ranging from 0.42 – 0.53). 

\begin{table}[ht]
\caption{Phase one classification report for ML algorithms.}\label{tab2}
\begin{tabular*}{\textwidth}{@{\extracolsep{\fill}}lllll}
\toprule
ML Algorithm        & Accuracy & Precision & Recall & F-1           \\
\midrule
Decision trees      & 0.34     & 0.38      & 0.34   & 0.34          \\
Naïve Bayes         & 0.43     & 0.53      & 0.43   & 0.45          \\
k-NN                & 0.39     & 0.52      & 0.39   & 0.42          \\
Logistic regression & 0.53     & 0.56      & 0.53   & \textbf{0.53} \\
AdaBoost            & 0.07     & 0.09      & 0.07   & 0.05          \\
SGD                 & 0.51     & 0.59      & 0.51   & 0.51          \\
SVM                 & 0.47     & 0.45      & 0.47   & 0.42          \\
Ensemble learning   & 0.48     & 0.43      & 0.48   & 0.42          \\
\bottomrule
\end{tabular*}
\end{table}

\noindent The second phase of the study involved training ML models for each of the twelve classes separately. In each iteration, the dataset was binary encoded, where 1 represented the documents of the class in consideration, while the rest were labelled as 0. Table~\ref{tab3} records the performance measures for selected eight classes out of the twelve classes the models were trained. The eight classes were picked to represent the classes with the least number of documents, the classes with the most and those with a moderate number of documents. For the selected classes  in Table~\ref{tab3}, the document spread is: Child Protection-26, Cybersecurity-40, Data Privacy-13, Data Systems-84, Digital Finance-149, Digital Inclusion-89, Digital Information Services- 59, and Policy \& Regulations-31. 

Similar to phase 1, since the dataset is class imbalanced, the F1-score was utilised to evaluate the model performance on the test dataset. As per the values recorded in Table~\ref{tab3}, the model performances were significantly better for most classes than fitting a single model for a multiclass problem. The first phase of the study achieved a maximum performance of 0.53 under logistic regression by fitting a single model for the test dataset. However, training separate models for each class using the OvR approach successfully resulted in better performances that reached up to 0.8 for some classes. Out of the best classifiers identified for the classes under the OvR approach, the least desirable performance of 0.4 was recorded for the Data Systems class under k-NN. Further supporting the conclusion of phase one, none of the classes recorded their best performance under decision trees and AdaBoost, while the best results were recorded for SGD and logistic regression. 

\begin{table}
\caption{Classification report for ML algorithms under OvR.}\label{tab3}
\begin{tabular*}{\textwidth}{@{\extracolsep{\fill}}llllllllll}
\toprule
\begin{tabular}[c]{@{}l@{}}ML\\ Algorithm\end{tabular} &
  \rotatebox{90}{\begin{tabular}[c]{@{}l@{}}Performance\\ Measures\end{tabular}} &
  \rotatebox{90}{\begin{tabular}[c]{@{}l@{}}Child\\ Protection\end{tabular}} &
  \rotatebox{90}{\begin{tabular}[c]{@{}l@{}}Cyber-\\ security\end{tabular}} &
  \rotatebox{90}{\begin{tabular}[c]{@{}l@{}}Data\\ Privacy\end{tabular}} &
  \rotatebox{90}{\begin{tabular}[c]{@{}l@{}}Data\\ Systems\end{tabular}} &
  \rotatebox{90}{\begin{tabular}[c]{@{}l@{}}Digital\\ Finance\end{tabular}} &
  \rotatebox{90}{\begin{tabular}[c]{@{}l@{}}Digital\\ Inclusion\end{tabular}} &
  \rotatebox{90}{\begin{tabular}[c]{@{}l@{}}Digital\\ Information\end{tabular}} &
  \rotatebox{90}{\begin{tabular}[c]{@{}l@{}}Policy \& \\ Regulations\end{tabular}} \\
\midrule
\multirow{4}{*}{Decision trees} &
  Accuracy &
  0.93 &
  0.94 &
  0.96 &
  0.82 &
  0.81 &
  0.74 &
  0.79 &
  0.79 \\
 & 
  Precision &
  0.42 &
  0.75 &
  0.44 &
  0.00 &
  0.75 &
  0.43 &
  0.19 &
  0.11 \\
 &
  Recall &
  0.45 &
  0.40 &
  0.57 &
  0.00 &
  0.54 &
  0.47 &
  0.23 &
  0.18 \\
 &
  F-1 &
  0.43 &
  0.52 &
  0.50 &
  0.00 &
  0.63 &
  0.45 &
  0.21 &
  0.14 \\
\multirow{4}{*}{Naïve Bayes} &
  Accuracy &
  0.81 &
  0.77 &
  0.92 &
  0.82 &
  0.70 &
  0.58 &
  0.85 &
  0.66 \\
 &
  Precision &
  0.23 &
  0.26 &
  0.30 &
  0.00 &
  0.50 &
  0.33 &
  0.41 &
  0.21 \\
 &
  Recall &
  1.00 &
  1.00 &
  0.86 &
  0.00 &
  1.00 &
  0.79 &
  0.64 &
  0.94 \\
 &
  F-1 &
  0.38 &
  0.41 &
  0.44 &
  0.00 &
  0.67 &
  \textbf{0.47} &
  0.50 &
  0.34 \\
\multirow{4}{*}{k-NN} &
  Accuracy &
  0.90 &
  0.95 &
  0.94 &
  0.68 &
  0.82 &
  0.49 &
  0.69 &
  0.72 \\
 &
  Precision &
  0.36 &
  0.67 &
  0.30 &
  0.30 &
  0.68 &
  0.23 &
  0.19 &
  0.11 \\
 &
  Recall &
  0.91 &
  0.80 &
  0.43 &
  0.61 &
  0.77 &
  0.53 &
  0.50 &
  0.29 \\
 &
  F-1 &
  0.51 &
  0.73 &
  0.35 &
  \textbf{0.40} &
  0.72 &
  0.33 &
  0.28 &
  0.16 \\
\multirow{4}{*}{Log. Reg} &
  Accuracy &
  0.95 &
  0.95 &
  0.98 &
  0.78 &
  0.86 &
  0.73 &
  0.86 &
  0.89 \\
 &
  Precision &
  0.53 &
  0.62 &
  0.80 &
  0.30 &
  0.71 &
  0.42 &
  0.44 &
  0.41 \\
 &
  Recall &
  0.82 &
  0.87 &
  0.57 &
  0.18 &
  0.91 &
  0.44 &
  0.50 &
  0.41 \\
 &
  F-1 &
  0.64 &
  0.72 &
  0.67 &
  0.23 &
  \textbf{0.80} &
  0.43 &
  0.47 &
  \textbf{0.41} \\
\multirow{4}{*}{AdaBoost} &
  Accuracy &
  0.95 &
  0.96 &
  0.96 &
  0.78 &
  0.81 &
  0.75 &
  0.85 &
  0.86 \\
 &
  Precision &
  0.62 &
  0.89 &
  0.44 &
  0.36 &
  0.76 &
  0.44 &
  0.14 &
  0.17 \\
 &
  Recall &
  0.45 &
  0.53 &
  0.57 &
  0.27 &
  0.55 &
  0.37 &
  0.05 &
  0.12 \\
 &
  F-1 &
  0.53 &
  0.67 &
  0.50 &
  0.31 &
  0.64 &
  0.41 &
  0.07 &
  0.14 \\
\multirow{4}{*}{SGD} &
  Accuracy &
  0.98 &
  0.96 &
  0.97 &
  0.79 &
  0.84 &
  0.77 &
  0.88 &
  0.88 \\
 &
  Precision &
  0.9 &
  0.77 &
  0.62 &
  0.39 &
  0.71 &
  0.54 &
  0.47 &
  0.14 \\
 &
  Recall &
  0.82 &
  0.67 &
  0.71 &
  0.27 &
  0.82 &
  0.16 &
  0.41 &
  0.06 \\
 &
  F-1 &
  \textbf{0.86} &
  0.71 &
  \textbf{0.67} &
  0.32 &
  0.76 &
  0.25 &
  0.44 &
  0.08 \\
\multirow{4}{*}{SVC} &
  Accuracy &
  0.97 &
  0.96 &
  0.96 &
  0.83 &
  0.84 &
  0.75 &
  0.89 &
  0.90 \\
 &
  Precision &
  1.00 &
  0.79 &
  0.00 &
  0.67 &
  0.68 &
  0.46 &
  0.67 &
  0.25 \\
 &
  Recall &
  0.55 &
  0.73 &
  0.00 &
  0.06 &
  0.89 &
  0.40 &
  0.18 &
  0.06 \\
 &
  F-1 &
  0.71 &
  \textbf{0.76} &
  0.00 &
  0.11 &
  0.78 &
  0.42 &
  0.29 &
  0.10 \\
\multirow{4}{*}{Ensemble learning} &
  Accuracy &
  0.97 &
  0.97 &
  0.97 &
  0.82 &
  0.82 &
  0.75 &
  0.85 &
  0.89 \\
 &
  Precision &
  1.00 &
  1.00 &
  0.75 &
  0.43 &
  0.92 &
  0.46 &
  0.41 &
  0.33 \\
 &
  Recall &
  0.55 &
  0.60 &
  0.43 &
  0.09 &
  0.43 &
  0.40 &
  \textbf{0.64} &
  0.24 \\
 &
  F-1 &
  0.71 &
  0.75 &
  0.55 &
  0.15 &
  0.59 &
  0.42 &
  0.50 &
  0.28 \\
\bottomrule
\end{tabular*}
\end{table}

The most noteworthy observation of this approach is the presence of clear evidence to prove that it is not always the same classification algorithm that works well for all the classes. For example, though the SGD model generated the highest performance for the Child Protection class with an F1-score of 0.86, the algorithm underperformed for the Digital Inclusion class, recording an F1-score of 0.25. Thus, the OvR approach allows us to train different classification algorithms for each of the twelve classes and optimise the classification results. This statement is further supported by the F1 scores recorded in Table~\ref{tab4} for each class under the logistic regression model trained during the first phase (multi-class classification). Though fitting the logistic regression model on test data generated a weighted average F1-score of 0.53, it performed poorly for some classes. The logistic regression model has performed well in predicting Child Protection and Digital Finance documents, while the performance for the Data Privacy class is nil. Nevertheless, adopting the OvR approach in the Data Privacy class made it possible to report an F1 score of 0.67 under the SGD model.

\begin{table}
\caption{Phase one classification report for logistic regression.}\label{tab4}
\begin{tabular*}{\textwidth}{@{\extracolsep{\fill}}llll@{}}
\toprule
Intervention area     & Precision & Recall & F-1  \\
\midrule
Child Protection      & 1.00      & 0.64   & 0.78 \\
Cybersecurity         & 0.60      & 0.75   & 0.67 \\
Data Privacy          & 0.00      & 0.00   & 0.00 \\
Data Systems          & 0.46      & 0.43   & 0.44 \\
Digital Finance       & 0.71      & 0.91   & 0.80 \\
Digital Inclusion     & 0.50      & 0.55   & 0.52 \\
Digital Information   & 0.23      & 0.58   & 0.33 \\
Policy \& Regulations & 0.30      & 0.32   & 0.33 \\
Accuracy              &           &        & 0.53 \\
Weighted Avg F-1      & 0.56      & 0.53   & 0.53\\
\bottomrule
\end{tabular*}
\end{table}

\noindent The eight classes in Table~\ref{tab3} have documents ranging from 13-149. Data Privacy is the class with the lowest number of documents, which is 13. However, it still managed to report a moderate performance of 0.67 under both logistic regression and SGD. Saying so, the Data Systems class did not meet an acceptable performance (F1-score of 0.4 under k-NN), given that it has 84 documents. The Digital Finance class, with the highest number of documents, met a significant performance of 0.8 for logistic regression. In contrast, with just 26 documents, Child Protection recorded the highest performance of 0.86 under the SGD model. This brings us to the point that it is not always the number of training samples available for each class that determines the model's performance. 

Facts like dissimilarity among classes and similarity within classes also affect the performance of a model. In agreement, Child Protection and Digital Finance classes have several terms unique to the class. For example, words like “child”, “sexual”, and “adolescent” are unique for the Child Protection class (Fig.~\ref{Fig-4}). But for a class like Digital Inclusion, the frequent words indicated in Fig.~\ref{Fig-4} are network, people, ict and phone. Given that the entire dataset is centred around digital development efforts, those words can be common to all the classes, resulting in weak performances for the models trained for such classes. With the recorded results, it can be stated that adapting an OvR approach generates more significant results, at least for some classes, than fitting a single model for all.

\section{Conclusion}
Automated text document classification is commonly used in various industries like customer service, news media, legal, healthcare, e-commerce, etc., to facilitate efficient information retrieval, aid decision making and enhance overall end-user satisfaction. With the rising number of digital development initiatives, a vast data pool with reports on such initiatives is evolving. This study has utilised several ML algorithms to automate the classification process of digital development reports, which will help world organisations such as the World Bank to effectively track and monitor development initiatives worldwide. 

In the study's first phase, a single model was fitted for a multiclass dataset. However, the weighted average F1-score did not report beyond 0.53 due to insufficient documents for some intervention areas and similarities among classes. This is a drawback of the USAID DEEM dataset being used for the study. As nations have started to embrace digital development lately, USAID continues to add new reports to the database. Therefore, future performances can be optimised by augmenting the dataset with reports from underrepresented intervention areas.

To overcome the drawbacks of the first phase of the study, an OvR approach was adopted, and comparatively, it yielded better performance for the custom dataset. The improved performance is due to models being fitted for a binary classification task, i.e., the models were trained to distinguish between two classes as a document belonging to the class or not. On the other hand, in the multiclass classification approach, a model was trained to classify a document into one of the twelve classes, making the process complex as the models must learn variation among twelve classes. 

Accordingly, for a multiclass classification task where training data points are limited, following an OvR approach will result in better performances. So, the recommended method for this study is to train separate models for each of the twelve classes and finally call them into a single function that can output all the classes a document belongs to, along with a probability score. Under this approach, unlike training a single model for multiclass classification, which maps one class for a document, the model will map more than one class for a document due to training errors. The above-stated issue is no harm to the USAID DEEM dataset, as it is more of a multi-label dataset. However, if one is dealing with a situation where a document can belong to only one of the classes, the class that most fits the document can be chosen based on the highest probability score. 

For simplicity, if one is interested in training a single ML model for multiclass classification, the recommended algorithm is SGD. The conclusion is drawn based on the average of F1 scores reported across the twelve intervention areas in phase 2. In contrast, other algorithms performed exceptionally well for certain classes and reported very weak performances for the rest of the classes. Here, one might argue that the results reported under phase one indicated logistic regression as the best-performing model for multiclass classification. Though the particular model reported a higher average performance, it reported a null F1-score for the Data Privacy class. This is in contrast to SGD, which performed moderately for all the classes with a weighted average F1-score of 0.51.

However, it should be noted that no significant performance was yielded by either the OvR or the full model for some of the classes. Few classes remained underperforming due to the lack of data points and the limited differences between classes. To elevate the overall outcome of the research, future work should address training models on a larger dataset while considering the possibility of overfitting. As digital development keeps rising, so will the reports on such initiatives. Therefore, there is space to make the dataset sounder and extend the study to include development reports pooled by other organisations. This is a positive sign to integrate DL models in future work, which were found to be outperforming ML algorithms for large datasets~\cite{behera2019performance}. 

\bibliographystyle{splncs04}
\bibliography{references}
\end{document}